\documentclass[a4paper,twocolumn]{article}
\usepackage[english]{babel}
\usepackage[nottoc]{tocbibind}
\usepackage{jsaiac}
\DeclareUnicodeCharacter{2212}{-}

\usepackage{breakcites}
\usepackage{makecell}
\usepackage{multirow}
\usepackage{amsmath}
\newcommand*\quot[2]{{^{\textstyle #1}\big/_{\textstyle #2}}}
\usepackage{mathtools}
\usepackage{diagbox}
\usepackage{caption}
\usepackage{subcaption}
\usepackage{graphicx}
\usepackage{enumitem}
\usepackage{xcolor}
\usepackage{adjustbox}
\usepackage[thinlines]{easytable}
\usepackage{cleveref}
\usepackage{algorithm}
\usepackage{algpseudocode} 
\usepackage{cancel,soul}

\newcommand{\eg}{\textit{e.g.}}

\newcommand{\dg}{\textit{Dialogue-Graph}}
\newcommand{\pjds}{\textit{PLATO-JDS}}
\usepackage{xcolor}
\usepackage{soul}
\definecolor{bluenode}{HTML}{00a7db}
\definecolor{rednode}{HTML}{ea4e00}
\definecolor{purplenode}{HTML}{9d00d7}
\definecolor{graynode}{HTML}{898989}
\definecolor{ForestGreen}{RGB}{34,139,34}
\definecolor{cgreennode}{HTML}{006B3C}

\makeatletter
\newcommand{\thickhline}{%
    \noalign {\ifnum 0=`}\fi \hrule height 1pt
    \futurelet \reserved@a \@xhline
}
\newcolumntype{"}{@{\hskip\tabcolsep\vrule width 1pt\hskip\tabcolsep}}
\makeatother

\usepackage[whole]{bxcjkjatype}
\usepackage{times}
\usepackage{latexsym}
\usepackage[T1]{fontenc}
\usepackage[utf8]{inputenc}
\usepackage{microtype}

\title{Topic-switch adapted Japanese Dialogue System based on PLATO-2}

\address{KDDI Research, Inc., AI Division, 356-0003 Saitama, Fujimino, Ohara, 2 Chome−1−15, 049-278-7441, do-zeng@kddi-research.jp.}

\author{Donghuo Zeng\first
\and
Jianming Wu\first
\and
Yanan Wang\first
\and
Kazunori Matsumoto
\and
Gen Hattori
\and
Kazushi Ikeda
}
\affiliate{
\first{KDDI Research, Inc. }
}

\begin{abstract}
Large-scale open-domain dialogue systems such as PLATO-2 have achieved state-of-the-art scores in both English and Chinese. However, little work explores whether such dialogue systems also work well in the Japanese language. In this work, we create a large-scale Japanese dialogue dataset, Dialogue-Graph, which contains 1.656 million dialogue data in a tree structure from News, TV subtitles, and Wikipedia corpus. Then, we train PLATO-2 using Dialogue-Graph to build a large-scale Japanese dialogue system, PLATO-JDS. In addition, to improve the PLATO-JDS in the topic switch issue, we introduce a topic-switch algorithm composed of a topic discriminator to switch to a new topic when user input differs from the previous topic. We evaluate the user experience by using our model with respect to four metrics, namely, coherence, informativeness, engagingness, and humanness. As a result, our proposed PLATO-JDS achieves an average score of \textbf{1.500} for the human evaluation with human-bot chat strategy, which is close to the maximum score of 2.000 and suggests the high-quality dialogue generation capability of PLATO-2 in Japanese. Furthermore, our proposed topic-switch algorithm achieves an average score of \textbf{1.767} and outperforms PLATO-JDS by \textbf{0.267}, indicating its effectiveness in improving the user experience of our system.
\end{abstract}

\def\BibTeX{{\rm B\kern-.05em{\sc i\kern-.025em b}\kern-.08em%
 T\kern-.1667em\lower.7ex\hbox{E}\kern-.125emX}}
\def\JBibTeX{\leavevmode\lower .6ex\hbox{J}\kern-0.15em\BibTeX}
\def\LaTeXe{\LaTeX\kern.15em2$_{\textstyle\varepsilon}$}

\begin{document}
\maketitle

\begin{figure*}[t!]
    \centering
    \includegraphics[width=0.8\textwidth]{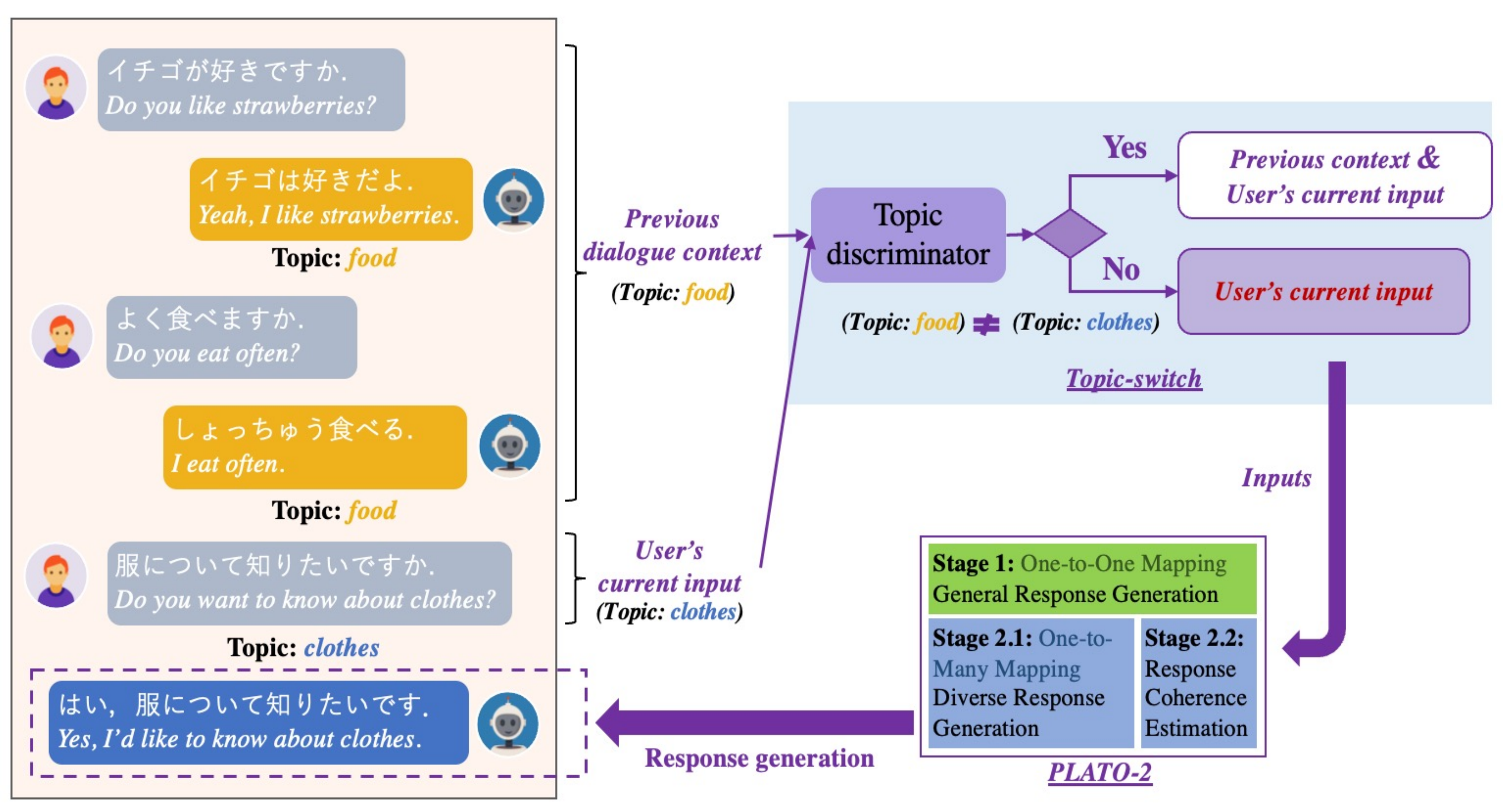}
    \caption{Overview of the topic-switch based PLATO-JDS. Given a user input with previous dialogue, the topic-switch consists of a topic discriminator that predicts if the user input is the same topic as previous dialogue, and switch to a new topic when it differs form previous dialogue's. The PLATO-JDS is a pretrained Japanese dialogue generation model for generating a response by giving specific inputs from the topic-switch.}
    \label{fig:fig1}
\end{figure*}

\section{Introduction}
Transformer-based methods are becoming fundamental techniques to developing human-like chatbots~\cite{devlin2018bert, gpt-3, golovanov2019large, wolf2019transfertransfo, zhou2021eva, shalaby2020building, adiwardana2020towards, radford2019language}. Large-scale open-domain dialogue systems such as PLATO-2~\cite{bao2020plato} is designed to scale the model up to billion parameters leading to high-quality open-domain chatbots, have achieved state-of-the-art scores in both English and Chinese. However, such large-scale dialogue systems are rarely explored in Japanese.

In this work, we create a large-scale Japanese dialogue dataset, Dialogue-Graph by collecting 1.656 million dialogue data in a tree structure from News, TV subtitles and Wikipedia corpus. Then, we train PLATO-2 using Dialogue-Graph to build a large-scale Japanese dialogue system, PLATO-JDS. Moreover, we study some cases generated by PLATO-JDS and the result suggests that PLATO-JDS is difficult to suitably switch to a new topic during a dialogue. To solve this issue to further improve the user experience of our system, we introduce a topic-switch algorithm composed of a topic discriminator to switch previous topic to a new topic when user input differs from previous topic. The topic discriminator is a BERT-based binary classifier~\cite{devlin2018bert} that used to predict whether the user input belongs to previous topic or not. The topic discriminator is a BERT-based binary classifier~\cite{devlin2018bert} used to predict whether the user input belongs to previous topic or not. As shown in Figure~\ref{fig:fig1}, if the result is "Yes", the concatenation of the user input and previous dialogue is used as the input to PLATO-JDS, otherwise, only the user input is used.

We evaluate PLATO-JDS in terms of four metrics, namely, coherence, informativeness, engagingness and humanness~\cite{bao2020plato}, and the results demonstrate that our proposed PLATO-JDS achieves scores of 1.600, 1.467, 1.467, and 1.467 on four metrics, respectively for human evaluation with the human-bot chat strategy, and the average score is \textbf{1.500} and close to the maximum score of 2.000, suggesting that PLATO-2 can be adopted to achieve high-quality dialogue generation in Japanese. Furthermore, our proposed topic-switch algorithm achieves the average score of \textbf{1.767} and outperforms PLATO-JDS by \textbf{0.267}, indicating its effectiveness in improving the user experience of our system.

Our contributions include: 1) we manually collected 1.656 million dialogue data to build a tree-structured large Japanese dataset \dg; 2) we trained a Japanese PLATO-2, PLATO-JDS by using \dg. Both human and automatic evaluations demonstrate that PLATO-JDS is effective to achieve high-quality dialogue generation; 3) Moreover, we introduced a topic-switch algorithm to further improve the user experience of PLATO-JDS.

\section{Related Work} \label{relatedwork}
Dialogue systems in the Japanese language domain have been developed by employing basic deep learning techniques, such as the AI love counseling system~\cite{docomonttmydaiz}. There are also some chatbot applications, such as Clova ~\cite{shin2019end} and Rinna ~\cite{wu2016} that employ state-of-the-art NLP methods to improve the performance of dialogue generation. The Japanese language model~\cite{rinna_pretrained2021} has trained two types of models on a public corpus, including the GPT-2~\cite{radford2019language} and RoBERTa~\cite{liu2019roberta} models. Considering a large-scale dataset can be used to train high-performance dialogue systems and should be taken into account. Current works~{~\cite{sugiyama2021empirical, rikters2019}} try to collect a large-scale dataset to improve Japanese dialogue system. Even so, few large-scale Japanese conversation systems have been trained by using a large Japanese language dataset, such as those in the millions in terms of size.

Recently, large-scale transformer-based dialogue generation models have significantly improved the performance of open-domain chatbots~\cite{zhang2019dialogpt, bao2019plato, bao2020plato, bao2021plato, rolleretal2021recipes, lewis2019bart}. DialoGPT~\cite{zhang2019dialogpt} is a transformer-based dialogue generation model trained by using 147M dialogue-like exchanges from Reddit comments and shows good performance in generating context-based responses by giving a single user input. PLATO ~\cite{bao2019plato} focuses on performing one-to-many dialogue generation~\cite{kim2020sequential} (\eg, given a single user input, the model can reply with more than one response) that is able to improve DialoGPT. Furthermore, PLATO-2~\cite{bao2020plato} scales the PLATO model up to billions of parameters to achieve a new state-of-the-art open-domain dialogue systems by introducing curriculum learning~\cite{curriculumLearning}, which contains pretraining on a one-to-one generation subtask and finetuning on a one-to-many generation subtask. However, PLATO-2 is evaluated on English and Chinese data, and there is no work exploring its effectiveness on Japanese data to facilitate the development of Japanese dialogue systems. To construct a high-quality dialogue system, control generation methods~\cite{dathathri2019plug, keskar2019ctrl, madotto2020plug, smith2020controlling, du2021sidecontrol} have been widely applied in dialogue system. In this study, PLATO-JDS is trained by curriculum learning, which controls coherent and fluent responses through two steps, as shown in Figure~\ref{fig:plato-2-overview}. 

To improve the user experience of dialogue systems, it is necessary to make advanced models have very good topic adaptability. The work~\cite{xu2021topic} extracts topic segments from dialogue history in an unsupervised way to select the most appropriate response for the topic-switch as needed. The work~\cite{xia2022dialogue} also adopts the topic segments method but also learns inter-segment relationships to improve the topic-switch. The work~\cite{sugiyama2021empirical} collect a mixed dataset with three characteristics: personality, empathy, and knowledge, to fine-tune BlenderBot to conduct a chit-chat system, which can switch topics frequently. Although these models track global topic flow throughout multi-turn dialogue, they have difficulty eliminating the interference of historical dialogues when users suddenly switch topic that is not related to the previous dialogue. Instead of inputting all the historical dialogues and user's input as previous methods, we propose a topic-switch algorithm with a topic discriminator to determine what the inputs of the trained PLATO-JDS are.

\section{Dataset} \label{datasets}
In this section, we describe the details of Japanese dataset \dg~in \S \ref{datacollection} and the data preprocessing in \S \ref{preprocessing}.
\begin{figure}[h]
    \centering
    \includegraphics[width=\linewidth]{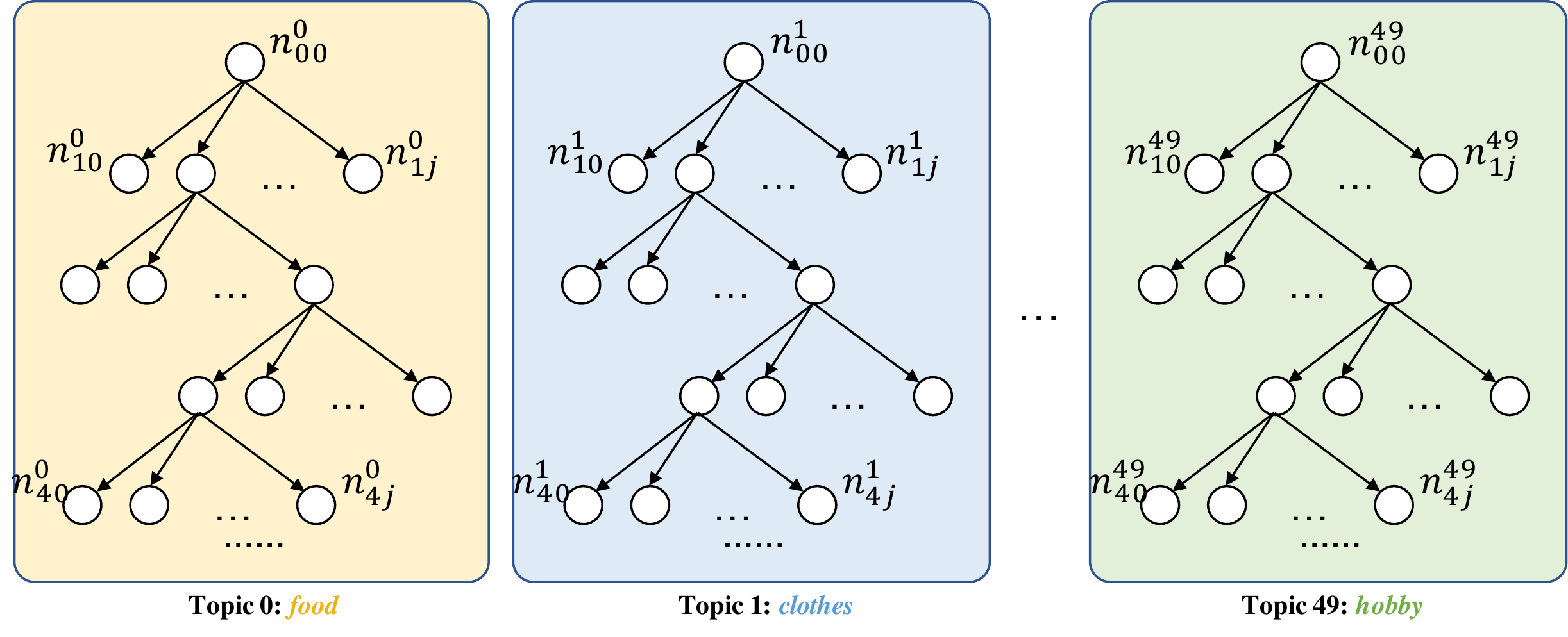}
    \caption{\dg: Start from \textbf{topic 0:food} with an initial utterance ${n}^{0}_{0,0}$:"What is your favorite food?", the \dg~ grows to a one-to-many dialogue data containing eight dialogue turns in max. Here, one-to-many indicates that one user input can accommodate many responses.}
    \label{fig:dialogueGraph}
\end{figure}
\subsection{Data collection} \label{datacollection}
In this work, we use the crowd-driven system~\cite{ikeda2018utilizing} to collect our dialogue data. This system can greatly improve the efficiency of dialogue data collection because it uses an asynchronous approach to dialogue creation, where workers can create an utterance at any time without waiting for the previous workers to finish creating the utterances. Specifically, when a worker starts using the system, the system will assign a dialogue that requires the workers' input, and the worker creates an utterance based on the given topic as well as the previous utterances.

We recruit 60 participants to collect the dialogue dataset. We assumed that a dialogue consists of alternating utterances between role A and role B, all participants were divided into two groups of 30 individuals playing role A and role B. One role can create utterances for multiple dialogues, and a dialogue is created by multiple workers. In the end, we select only high-quality conversations by evaluation method the same as~\cite{ikeda2018utilizing} and remove those dialogues that are not common sense, politically sensitive, and ethical concerns. To increase the efficiency of data collection and allow participants to create dialogues faster, we set the system to be accessible from 12:00 am to 3:00 pm. The dialogue data collection took over three months, costing 4 yen per utterance input and 32 yen per dialogue (about \$0.30).

Inspired by the studies~\cite{galitsky2018building, cheng2020conversational}, we also create tree-structure dialogues, with 9 levels from the first head node to the leaf node, i.e. the length of the conversation is 8 turns. We use 50 topics and each topic has an initial utterance, the same as the work~\cite{ikeda2018utilizing},
for example, the topic is family and his initial utterance is "Do you live with someone?". A worker created 8 utterances based on the starting utterance,
which became a dialogue. Then, we evaluated the collected dialogues using the same evaluation method as in~\cite{ikeda2018utilizing}. This method leverages the quality of the dialogues from three perspectives: efficiency, quality, and workers' interest. After the data collection was completed, we hired the same 60 participants to do the evaluation.

In Figure~\ref{fig:dialogueGraph}, we define \dg~$\mathcal{G}^{Dia}=(\mathcal{N},\mathcal{E})$, $\mathcal{N}=\{{n}^{t}_{i,j}| 0 \leq t \leq 49, 0 \leq i \leq 8 , j \geq 0\}$, where $\mathcal{N}$ is the set of dialogue node ${n}^{t}_{i,j}$, $\mathcal{E}=\{{e}^{t}_{i,j\rightarrow m,n}|0 \leq t \leq 49, 0 \leq i,m \leq 8 , j,n \geq 0\}$ is the set of dialogue turns, such that ${e}^{0}_{0,0\rightarrow 1,0}$ represents a single utterance turn from  ${n}^{0}_{0,0}$ to ${n}^{0}_{1,0}$. Here, $t,i/m$ and $j/n$ indicate the number of topic types, utterance turns and candidates of the current dialogue node, respectively, the maximum number of $t$ and $i/m$ are $50$ and $8$, the minimum number of $j/n$ is $3$.    

\begin{figure}[t!]
    \centering
    \includegraphics[width=0.5\textwidth]{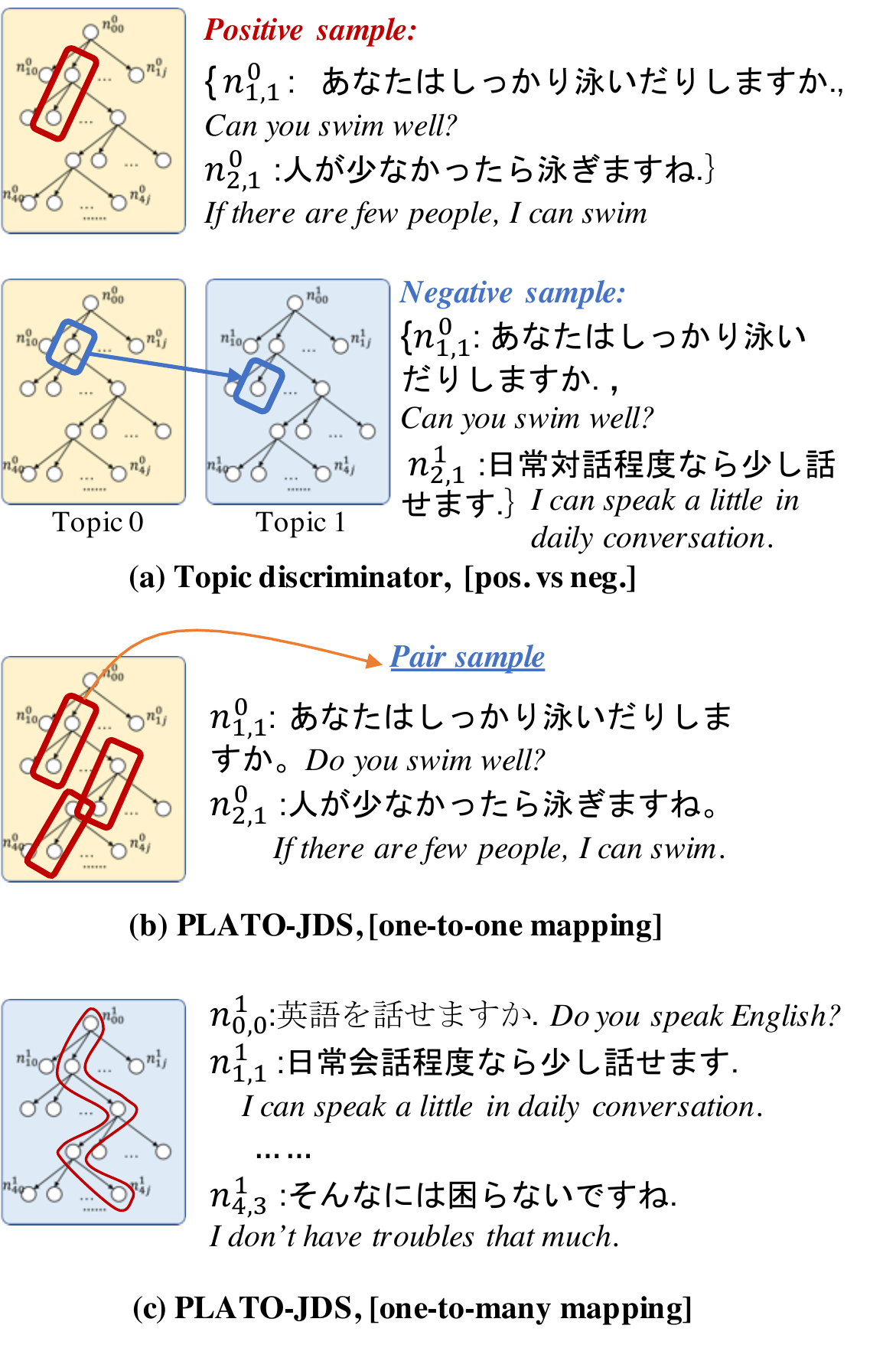}
    \caption{Training data samples for training a topic discriminator and \pjds.}
    \label{fig:1vs1}
\end{figure}

\subsection{Preprocessing} \label{preprocessing}

To prepare training data for training the topic discriminator and \pjds~, we follow three policies to preprocess data from \dg. We split each preprocessed dataset to training, validation and testing set with a common rate of $90\%:5\%:5\%$.
\begin{itemize}
\item \textbf{pos. vs neg.} As shown in Figure~\ref{fig:1vs1} (a), we extract arbitrary dialogue turns from the same topic and annotate them as positive samples. On the other hand, we annotate the counterpart as negative samples.
\item \textbf{one-to-one mapping.} We prepare training data for the stage 1 of \pjds. As shown in Figure~\ref{fig:1vs1} (b), we employ the breadth-first search method (BFS) to extract 1.68M data samples in total. BFS is designed to traverse dialogue nodes in the graph to obtain utterance pairs. Starting from the root node, we extract all utterance turns as training data (\eg, ${n}^{0}_{0,0}\rightarrow{n}^{0}_{1,0}$, ${n}^{0}_{0,0}\rightarrow{n}^{0}_{1,3}$). The definition of an utterance turn is provided in \S \ref{datasets}.
\item \textbf{one-to-many mapping.} We prepare training data for the stage 2 of \pjds. As shown in Figure~\ref{fig:1vs1} (c), we use the depth-first search method (DFS) method to generate 1.26M data samples. DFS is designed to traverse dialogue nodes in the graph to obtain all eight-turn dialogue as training data (\eg, ${n}^{0}_{0,0}\rightarrow{n}^{0}_{1,0}\hspace{-0.1cm}\rightarrow\hspace{-0.1cm}{n}^{0}_{2,1}\hspace{-0.1cm}\rightarrow\hspace{-0.1cm}{n}^{0}_{3,1}\hspace{-0.1cm}\rightarrow\hspace{-0.1cm}{n}^{0}_{4,1}\hspace{-0.1cm}\rightarrow\hspace{-0.1cm}{n}^{0}_{5,0}\hspace{-0.1cm}\rightarrow\hspace{-0.1cm}{n}^{0}_{6,0}\hspace{-0.1cm}\rightarrow\hspace{-0.1cm}{n}^{0}_{7,0}\hspace{-0.1cm}\rightarrow\hspace{-0.1cm}{n}^{0}_{8,2}$).
\end{itemize}

\section{Approach} \label{mymodel}
To achieve PLATO-JDS as shown in Figure~\ref{fig:fig1}, we design a topic-switch algorithm to utilize a topic discriminator to decide what the model inputs is by giving a user input and previous dialogue, then we feed it to a trained PLATO-JDS to generate a response. In this section, we first describe the detail of the topic discriminator in \S \ref{sec:topicdiscriminator} and PLATO-JDS in \S \ref{sec:plato-2} in detail, then we describe how the topic-switch algorithm works in \S \ref{sec:topicswitch}. 

\begin{figure}[t!]
    \centering
    \includegraphics[width=0.5\textwidth]{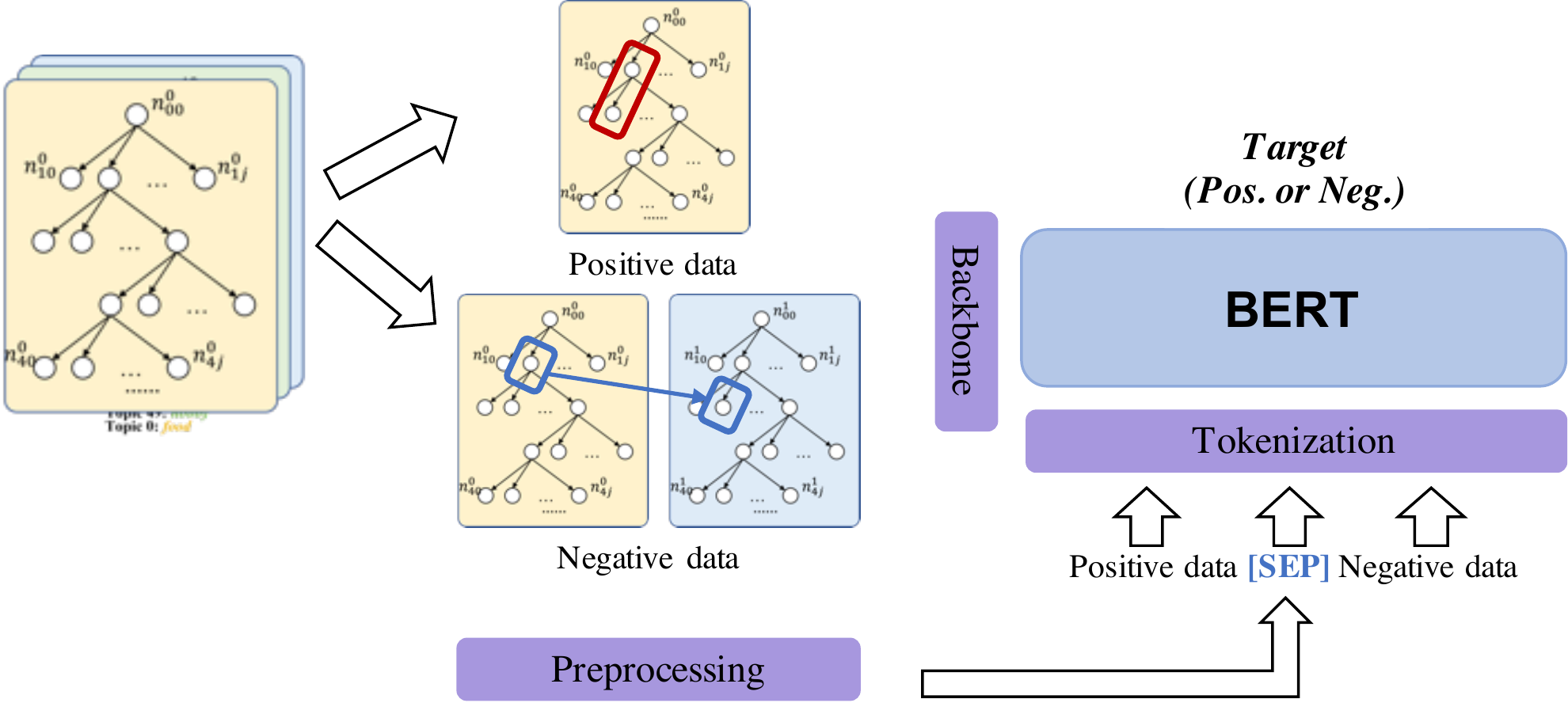}
    \caption{Overview of topic discriminator. It consists of data preprocessing described in \S ~\ref{preprocessing}, tokenization method employed to obtain word sequences, the backbone of BERT for training a binary classifier.}
    \label{fig:topicswitch}
\end{figure}

\subsection{Topic discriminator}\label{sec:topicdiscriminator}
The topic discriminator is a binary classifier that is designed to determine whether current user input belongs to a topic common to the previous dialogue. As shown in Figure~\ref{fig:topicswitch}, we build a BERT-based topic discriminator and train it by using the preprocessed Dialogue-Graph data described in \S \ref{preprocessing}. We annotate data that belongs to a common topic as positive samples and the counterpart as negative samples. We concatenate one turn containing an utterance pair as a single sentence to input into the tokenization method followed by the backbone of BERT. The objective of this training task is to discriminate whether two utterances in one pair belongs to a common topic by optimizing the binary cross entropy (BCE) loss as follows.
\begin{equation} 
{L_{BCE} = -\frac{1}{N} \sum_{i=1}^{N} (Y_{i} \cdot \log \hat{Y_{i}} +(1-Y_{i}) \cdot \log (1-\hat{Y_{i}}))}
\end{equation}
where $N$ is the number of samples, $Y_{i}$ denotes the ground-truth label (0 or 1), while $\hat{Y_{i}}$ is the predicted probability of the topic discriminator.

Here, we employ the SentencePiece~\cite{kudo2018sentencepiece} tokenization method to segment a raw input sentence directly into word sequences. SentencePiece is a language-agnostic tokenization method based on a subword idea to avoid unknown tokens while reducing the number of tokens. The Byte Pair Encoding (BPE)~\cite{gage1994new, sennrich2015neural} algorithm is a subword division algorithm employed by SentencePiece to shortlist high-frequency words as tokens and divide low-frequency words into two or more subwords as tokens. We use SentencePiece to split the Japanese dataset \dg~ into 48k tokens. Finally, we employ the same input representations as the original PLATO-2 model including token embedding, role embedding, turn embedding and position embedding, respectively. Here, token embedding is a pretrained embedding for different subwords. Role embedding is designed to distinguish the role of speakers during one dialogue. Turn embedding is assigned according to relative order when there are multi-turn dialogues in the conversion. Position embedding is obtained based on the token position in each dialogue.

\subsection{PLATO-JDS}\label{sec:plato-2}
We follow the original PLATO-2 model~\cite{bao2020plato} to train it on Japanese data to perform dialogue generation. PLATO-2 is a large-scale transformer based dialogue generation framework that is trained via curriculum learning~\cite{curriculumLearning}. Curriculum learning is a two-stage training strategy and is shown in Figure 3. In the first stage, a coarse-grained generation model is pretrained to learn response generation under different dialogue contexts by using the preprocessed one-to-one mapping data shown in Figure~\ref{fig:1vs1} (b). In the second stage, a fine-grained generative model and an evaluation model are trained by using the preprocessed one-to-many mapping data shown in Figure~\ref{fig:1vs1} (c). We adopt the same input representations as the topic discriminator model.

\textbf{Stage 1, General response generation}. One-to-one mapping is a conventional approach and it is an efficient way to learn high-level properties of response generation. Given a $i^{th}$ response $R(i)$ and its previous dialogue context $C(i)$, the model is trained by minimizing the negative log-likelihood (NLL) loss as follows,
\begin{equation}
    L_{NLL}^{General} = -~\textbf{E} \sum_{i=1}^{L}\log p(R(i)|C(i)) 
\end{equation}
where $L$ is the length of the generated response $R(i)$.

\textbf{Stage 2.1, Diverse Response Generation}. To further train the model to capture the relationship of one-to-many mapping, a discrete latent variable $z$ is introduced. The model estimates the latent distribution of training samples $p(z|C(i), R(i))$ and then generates sampled latent variable $p(R(i)|C(i), z)$. The $NLL$ loss of diverse response generation and the bag-of-words (BOW) loss are defined as follows.
\begin{equation}
    \begin{split}
        \begin{aligned}
    L_{NLL}^{Diverse} &=-\textbf{E}_{z\sim p(z|C(i), R(i))}\sum_{i=1}^{L}\log p(R(i)|C(i), z) \\[5pt]
    L_{BOW}^{Diverse} &=-\textbf{E}_{z\sim p(z|C(i), R(i))}\sum_{i=1}^{L}\log p(R(i)|C(i), z) \\[5pt]
    L^{Diverse} &=L_{NLL}^{Diverse}+L_{BOW}^{Diverse}
        \end{aligned}
    \end{split}
\end{equation}

where $L$ is the length of generated response $R(i)$. The final generation model is optimized by the above integrated loss $L^{Diverse}$.

\textbf{Stage 2.2, Coherent Response Selection}. Once diverse responses generated from one-to-many generation, the highest quality response is selected as the final output. The selection model is trained by estimating the coherence between a given utterance and its response by leveraging the capacity of distributed representation of the masked language model (MLM). Two objective loss of response coherence estimation (RCE) and MLM are as follows.

\begin{equation}
\begin{aligned}
L_{RCE}^{Selection} &= -\log p(l_{r} = 1|C(i), R(i))-\log p(l_{r} = 0|C(i), \hat{R(i)}) \\ \break
L_{MLM}^{Selection} &= -\textbf{E}\sum_{i\in L}\log p({x_{t}|\quot{x}{L}}) \\ \break
L^{Selection}&=L_{RCE}^{Selection}+L_{MLM}^{Selection}
\vspace{-2mm}
\end{aligned}
\end{equation}
where $l_{r}$ denote the generated response is consistent with dialogue text ($R(i)$) or not ($\hat{R(i)}$), $x$ denotes input tokens of context and response. $\{x_{i}\}_{i\in L}$ denotes mask tokens, and $\quot{x}{L}$ represents rest unmasked tokens.

\begin{figure}[t!]
    \centering
    \includegraphics[width=0.5\textwidth]{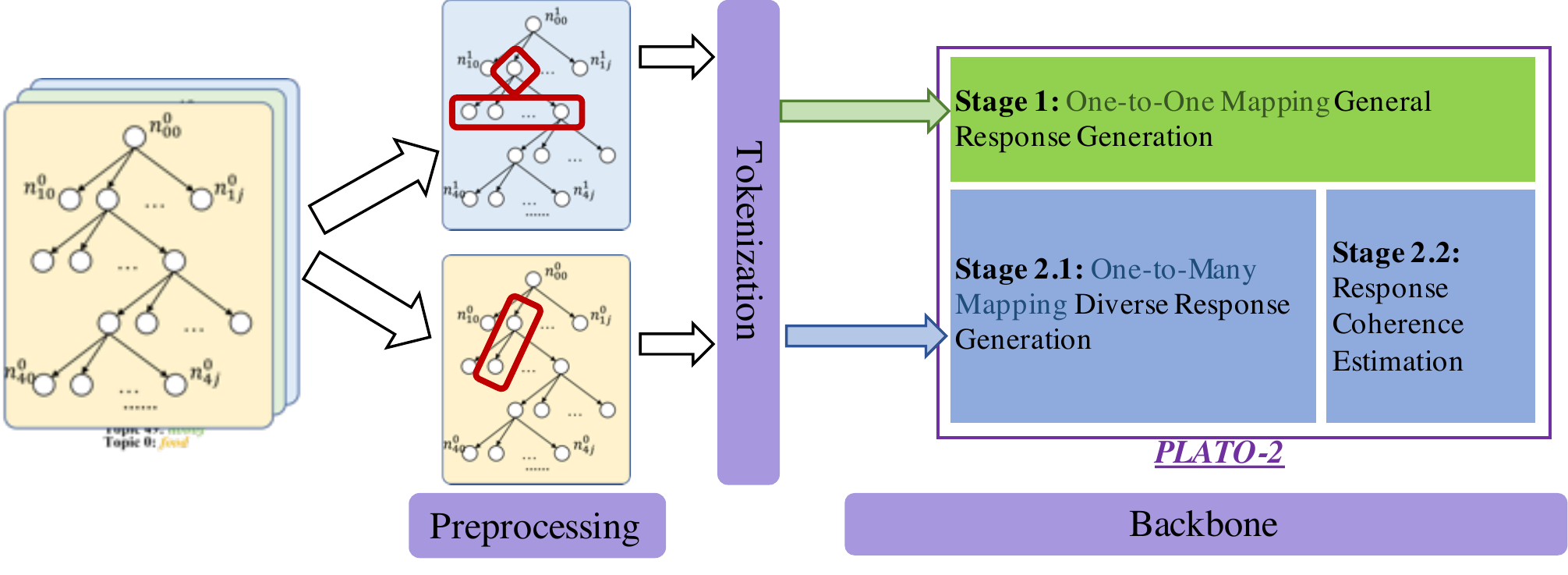}
    \caption{Overview of the PLATO-JDS model and its data processing. The data preprocessing described in \S ~\ref{preprocessing}, tokenization method as same as the topic discriminator (\S ~\ref{sec:topicdiscriminator}), and the backbone dialogue generation model of PLATO-2.}
    \label{fig:plato-2-overview}
\end{figure}

\subsection{Topic-switch algorithm}\label{sec:topicswitch}
A pretrained PLATO-JDS is used to generate a response by inputting the concatenation of the previous dialogue and current user input. We design a topic-switch algorithm to decide whether to include the previous dialogue used for response generation. We define the topic-switch algorithm in Algorithm~\ref{al:euclid}. We utilize a topic discriminator to predict whether the topic of current user input belongs to the previous dialogue topic by comparing with an experimental threshold. If we get a score that is less than the threshold, the current user input is probably unrelated to the previous dialogue and the input for the PLATO-JDS module will be automatically switched to current user input. Conversely, the input for the PLATO-JDS module will include previous dialogue. Here, we set the threshold at 0.61 based on the trade-off between the precision and recall, where the recall is 99.75\% and the precision is 95.72\%. 
\begin{algorithm}
  \caption{Topic-switch algorithm}\label{al:euclid}
   \textbf{Input}: $\varepsilon$: a threshold for deciding if switch to a new topic, $i$: the number of turns, $C(i)$: the previous dialogue of turn $i$, $\Phi (i)$: the user utterance of turn $i$, $\Gamma(a, b)$: the topic discriminator to predict whether $a$ and $b$ belong to a same topic. \\
   \textbf{Output}: the context of next utterance $C(i)$
  \begin{algorithmic}[1]
  \State $\beta = \Gamma(C(i),\Phi (i))$\hspace{-0.9cm} 
   \If{$\beta \le \varepsilon$}
    \State $C(i) \gets \Phi (i)$ \Comment{Switch to a new topic}
\ElsIf{$\beta > \varepsilon$}
    \State $C(i) \gets C(i) + \Phi (i)$ 
\EndIf
    \State $i \gets i + 1$ 
   \end{algorithmic}
\end{algorithm}

\section{Experiment}\label{experiment}
In this section, we discuss the evaluation metrics in \S \ref{metrics}. Then, we explain why we choose PLATO-2 as the backbone of dialogue generation module by discussing the result in the English data reported in PLATO-2's work in \S \ref{baseline}. To evaluate our proposed topic-switch based \pjds, we compare our model with \pjds~and report the comparison results in \S \ref{results}. Furthermore, we discuss the result of a case study in \S \ref{casestudy}, and the training detail in \S \ref{training}.

\subsection{Evaluation Metric}\label{metrics}
We evaluate our model from human and automatic perspectives. To achieve a fair comparison, we utilize the same evaluation metrics used in the PLATO-2's study~\cite{bao2020plato}. In terms of human evaluation, we asked 15 people \textit{(11 male and 4 female)} to perform a three-level evaluation \textit{(0:bad,1:neutral,2:good)} with respect to the utterance-level metrics of "coherence" and "informativeness", and the dialogue-level metrics of "engagingness" and "humanness" to evaluate the user experience. The definitions of these metrics are as follows:
\begin{itemize}
    \item Coherence: leveraging the response whether it is relevant to the current topic and consistent with the context.
    \item Informativeness: measuring the response whether it is informative and is appropriate.
    \item Engagingness: judging the dialogue whether the evaluator likes to talk with the speaker in a long dialogue.
    \item Humanness: assessing the dialogue whether the speaker is a human being or the response is natural.
\end{itemize}

In terms of automatic evaluation, to evaluate the model's capacity on lexical diversity, we employ the corpus-level metrics~\cite{li2015diversity} distinct-1 and distinct-2 \textit{(distinct-1/2 in Table~\ref{tab:selfchat},~\ref{tab:humanbot})}, which are defined as the number of distinct $unigrams$ and $bigrams$ as scaled by the total number of generated tokens in response generation.

\subsection{Baseline} \label{baseline}
We set PLATO-2 as our baseline since it is the state-of-the-art dialogue generation model on the English and Chinese domains. 
PLATO-2 has achieved the best performance in both human and automatic evaluations compared to the Blender model~\cite{rolleretal2021recipes} that mitigates undesirable toxic or bias traits of large corpora by introducing blended skills. However it requires extensive manual annotations. Considering the cost-performance ratio, we only implement PLATO-2 as \pjds~and compare it with our proposed topic-switch based \pjds~, both trained on \dg~in this work. We perform both evaluations through bot-bot and human-bot chat strategies and provide the detail in \S \ref{results}. The description of \pjds~is provided in \S \ref{sec:plato-2}.

\subsection{Training Details}\label{training}
The \pjds~ training include two stages.
In stage 1, we employed negative log-likelihood (NLL) loss to capture the general characteristics of response generation via one-to-one mapping learning. In stage 2.1, we took the sum of NLL loss and bag-of-words (BOW) loss to learn the fine-grained generation. Here, BOW loss is used to reflect the training process of discrete latent variables. In stage 2.2, the objective loss was the sum of losses of response coherence estimation and Masked Language Model (MLM) loss~\cite{bao2020plato}. More details of the above loss functions can be found in \S~\ref{sec:plato-2}. Moreover, we set the maximum sequence length of context and response to 512, and the size of position embedding to 256. We trained \pjds~for 25 days by using NVIDIA RTX A5000 GPU (24G x 2).

In addition, we show the loss curve of model training and validation process in the stages 1, 2.1, and 2.2 in Figure 7 in Appendices. It shows that overfitting does not occur during model training, and demonstrates that the trained model is reliable.

\subsection{Result}\label{results}

\subsubsection{Bot-bot chat strategy}\label{self_eval}
The bot-bot chat strategy is designed to evaluate dialogue generation by performing dialogue simulation between two chatbots to reduce time consumption and expense of full human evaluations, which require humans to spend time talking to the chatbot and scoring generated response. This strategy is commonly applied to achieve a fair evaluation of dialogue generation~\cite{lietal_2016_deep, bao2020plato}.
We randomly selected 200 questions from our dataset as the start to produce ten-turns dialogues per question for models. As the result shown in Table~\ref{tab:selfchat}, our topic-switch based \pjds~outperformed \pjds~ on all human evaluation metrics, which suggests that the responses generated by our topic-switch based \pjds~are coherent in the current dialogue context, and contains much more information than \pjds. In addition, from the scores of "Engagingness" and "Humanness" metrics, we believe that the topic-switch based \pjds~ is capable of improving the user experience on \pjds, which also proves the effectiveness of generating human-like responses. Furthermore, the result of distinct-1/2 demonstrated that our topic-switch based \pjds~contains a larger number of distinct $unigrams$ and $bigrams$ than \pjds, which indicates its capacity in terms of lexical diversity. In addition, from the average length of utterance and the average number of topics, our topic-switch based \pjds~ achieved higher point than \pjds, which further verifies the validity of the topic-switch on switching to a new topic depending on the user input.
\begin{table}[h]
\centering
\begin{adjustbox}{width=0.45\textwidth}
\begin{tabular}{r|cc}
\thickhline
\diagbox[width=7em]{\textbf{Metric}}{\textbf{Model}} & \textbf{w/o topic-switch} & \textbf{w/ topic-switch} \\ \hline
\multicolumn{3}{c}{\textbf{Human evaluation}}\\ \hline
Coherence & 1.667 & \textbf{1.800} \\
Informativeness & 1.533 &\textbf{1.867} \\
Engagingness & 1.533 &\textbf{1.733} \\
Humanness & 1.533 & \textbf{1.800} \\ \hline
Average score &1.567 &\textbf{1.800} \\ \hline
\multicolumn{3}{c}{\textbf{Automatic evaluation}}\\ \hline
Distinct-1/2 & 0.343/0.712 & \textbf{0.388/0.730} \\
\hline
Length (avg.) & 14.016 & \textbf{14.667} \\
Topics (avg.) & 4.6  & \textbf{6.8} \\
\thickhline
\end{tabular}
\end{adjustbox}
\caption{The bot-bot chat strategy: comparison results of \pjds~(w/o topic-switch) and topic-switch based \pjds~(w/ topic-switch).}
\label{tab:selfchat}
\end{table}
\vspace{-2mm}

\subsubsection{Human-bot chat strategy}\label{human_eval}
In addition to the bot-bot chat strategy, we also collected human-bot dialogue records to evaluate our model in a real-world setting.
We employed 15 participants to chat with the bot, and each participant was requested to talk to the bot for 50 turns. We generated a total of 750 dialogue data. The results are summarized in Table~\ref{tab:humanbot}. We obtained a very close comparison result as well as the result described in \S \ref{self_eval}. Although lower scores were obtained for our topic-switch based \pjds~compared with those in the bot-bot chat strategy, we achieved higher scores on all metrics compare to \pjds. 

To further confirm the effectiveness of the topic-switch, we also compared the number of topics on average in the generated dialogue. As shown in Tables~\ref{tab:selfchat} and~\ref{tab:humanbot}, topic-switch based \pjds~manages two more topics and longer dialogue than the \pjds, which demonstrates the effectiveness of the topic-switch on making \pjds~to generate a response based on a new topic. As a result, adapting the topic-switch algorithm that can improve \pjds~in both human and automatic evaluations and suggest the capability in improving the user experience. 
\begin{table}[h]
\centering
\begin{adjustbox}{width=0.45\textwidth}
\begin{tabular}{r|cc}
\thickhline
\diagbox[width=7em]{\textbf{Metric}}{\textbf{Model}} & \textbf{w/o topic-switch} & \textbf{w/ topic-switch}  \\ \hline
\multicolumn{3}{c}{\textbf{Human evaluation}}\\ \hline
Coherence & 1.600 & \textbf{1.733} \\
Informativeness & 1.467 &\textbf{1.800} \\
Engagingness & 1.467 &\textbf{1.800} \\
Humanness & 1.467 & \textbf{1.733} \\ \hline
Average score &1.500 &\textbf{1.767} \\ \hline
\multicolumn{3}{c}{\textbf{Automatic evaluation}}\\ \hline
Distinct-1/2 & 0.339/0.709 & \textbf{0.376/0.715} \\
\hline
Length & 13.970 & \textbf{14.333} \\
Topics (avg.) & 4.2  & \textbf{6.5} \\
\thickhline
\end{tabular}
\end{adjustbox}
\caption{The human-bot chat strategy: comparison results of \pjds~(w/o topic-switch) and topic-switch based \pjds~(w/ topic-switch).}
\label{tab:humanbot}
\end{table}

\subsection{Case study}\label{casestudy}
\begin{figure}[h]
    \centering
    \includegraphics[width=\linewidth]{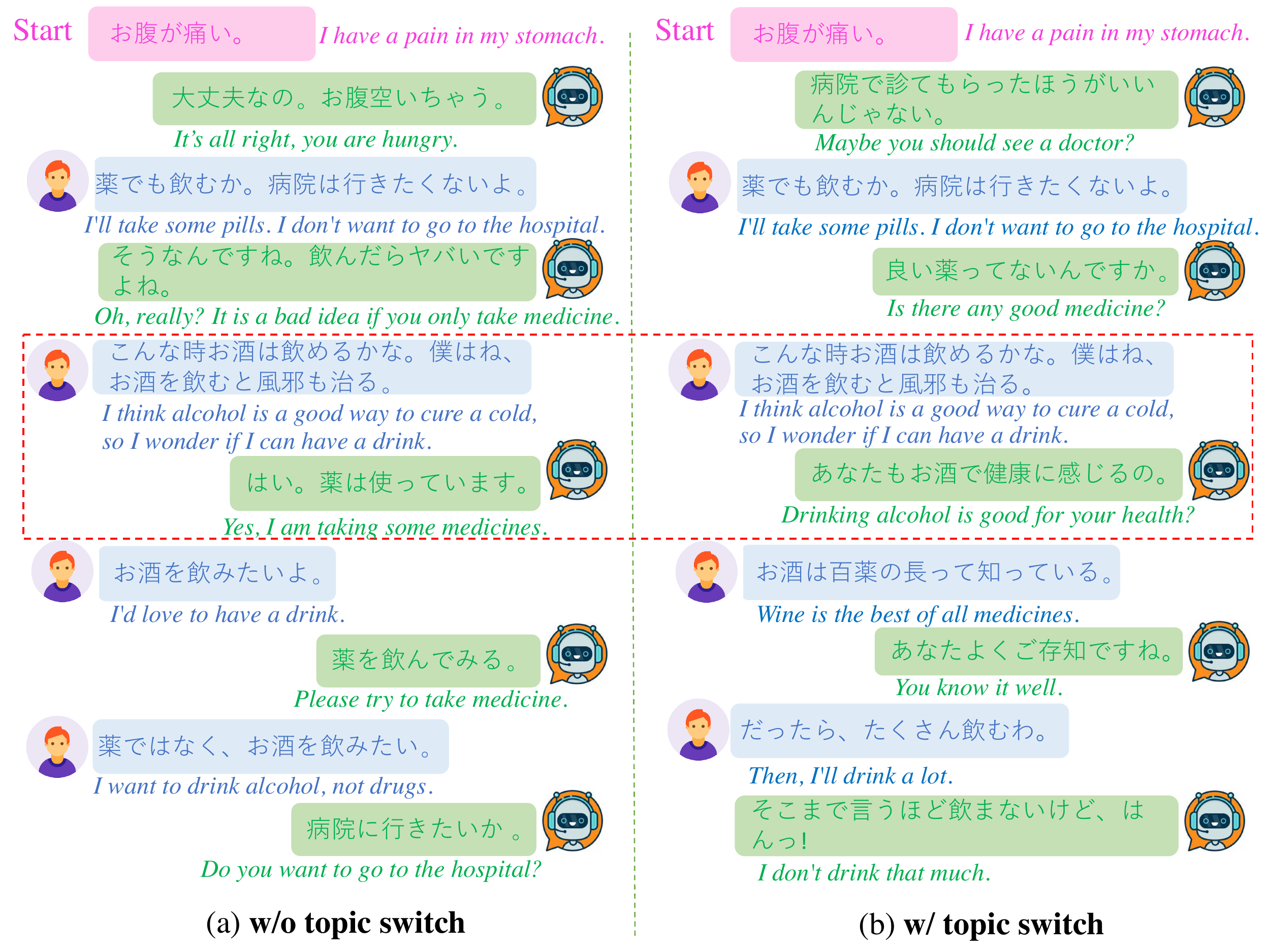}
    \caption{Two dialogue samples generated with the \pjds~with or without topic-switch models. Starting from the same utterance "I have a pain in my stomach" that belongs to the topic of "Health", when the user input change to the new topic of "Alcohol" in the red dotted box to chat; our  topic-switch based \pjds~ can switch to the "Alcohol" topic to generate an appropriate response but \pjds~ can not.}
    \label{fig:human_bot_samples}
\end{figure}

We pick up two human-bot dialogue samples of \pjds~with or without topic-switch by using a same start utterance in the same topic "beauty and health", which is to study how topic-switch affect the response generation. As shown in Figure~\ref{fig:human_bot_samples}, when user input changes to a new topic that differs from context in previous dialogue, we noted that our model with the topic-switch is able to generate an appropriate response but \pjds~can not. We believe that our proposed topic-switch forced \pjds~module to switch to a new topic to generate its relevant response, thereby demonstrating the effectiveness of the topic-switch. Furthermore, the case study also demonstrated that the coherence among all the evaluation results described in \S \ref{results}. 

\section{Conclusion}
\label{conclusion}
We built a PLATO-2 based Japanese dialogue system PLATO-JDS, and proposed a topic-switch algorithm to make PLATO-JDS switch to a new topic if user input differs from the previous topic. We built the largest Japanese dialogue generation dataset \dg~ by collecting 1.656M dialogue data from News, TV subtitles and Wikipedia corpus. 
our proposed PLATO-JDS achieved the average score of \textbf{1.500} for the human evaluation with human-bot chat strategy, and our proposed topic-switch algorithm further improved PLATO-JDS by \textbf{0.267} and achieved \textbf{1.767} human evaluation. The results suggest that PLATO-2 can be adopted to achieve high-quality dialogue generation in Japanese and the topic-switch is effective for improving the user experience of our system.

\bibliographystyle{apalike}
\bibliography{refs}

\appendix
\section*{Appendix}
\begin{figure}[h]
     \centering
     \begin{subfigure}{0.5\textwidth}
         \centering
         \includegraphics[width=\textwidth]{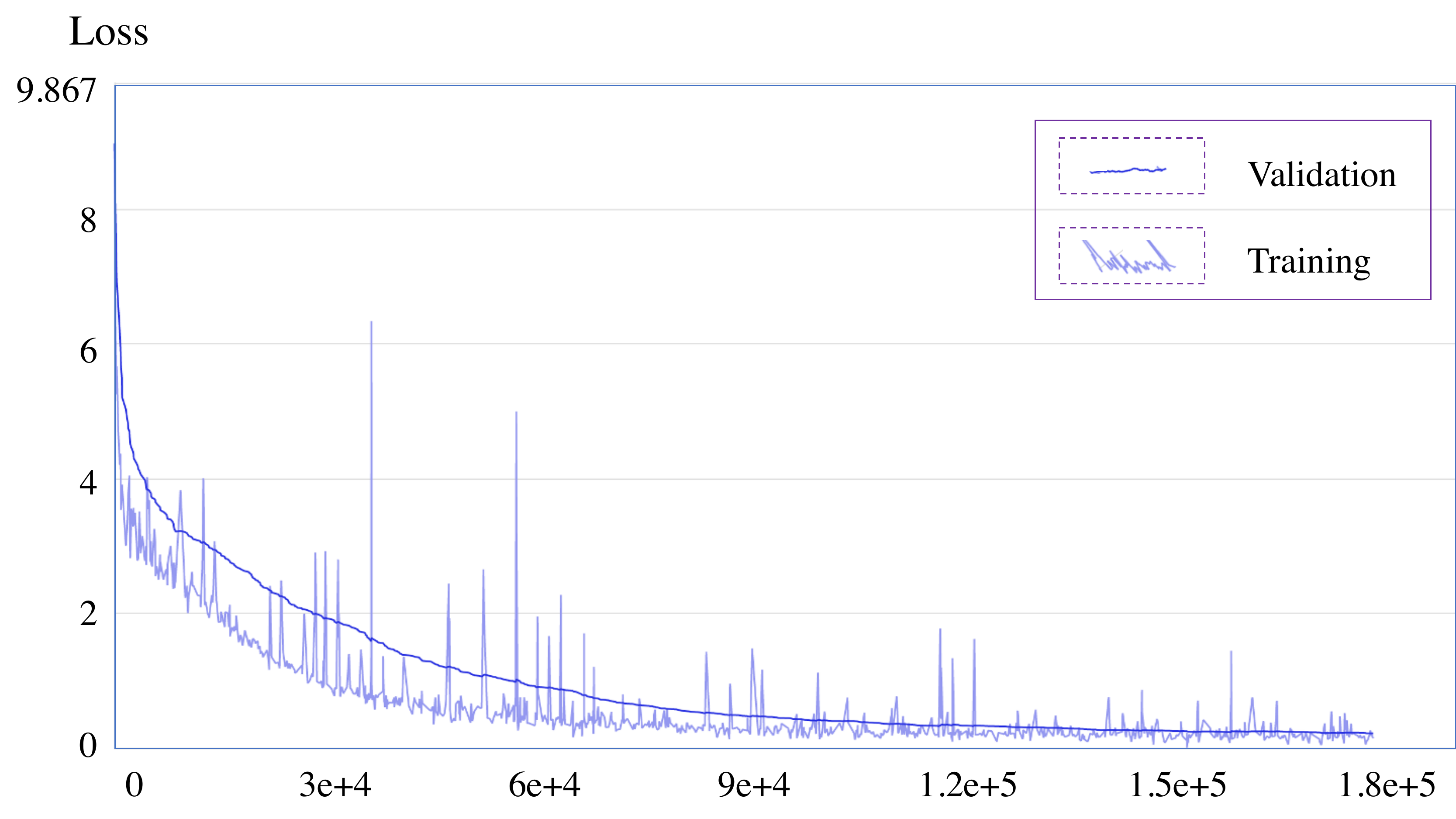}
         \caption{One to one mapping model training in the stage 1 of the \pjds~.}
         \label{fig:ppl_model1_one2one}
     \end{subfigure}
     \hfill
     \begin{subfigure}{0.5\textwidth}
         \centering
         \includegraphics[width=\textwidth]{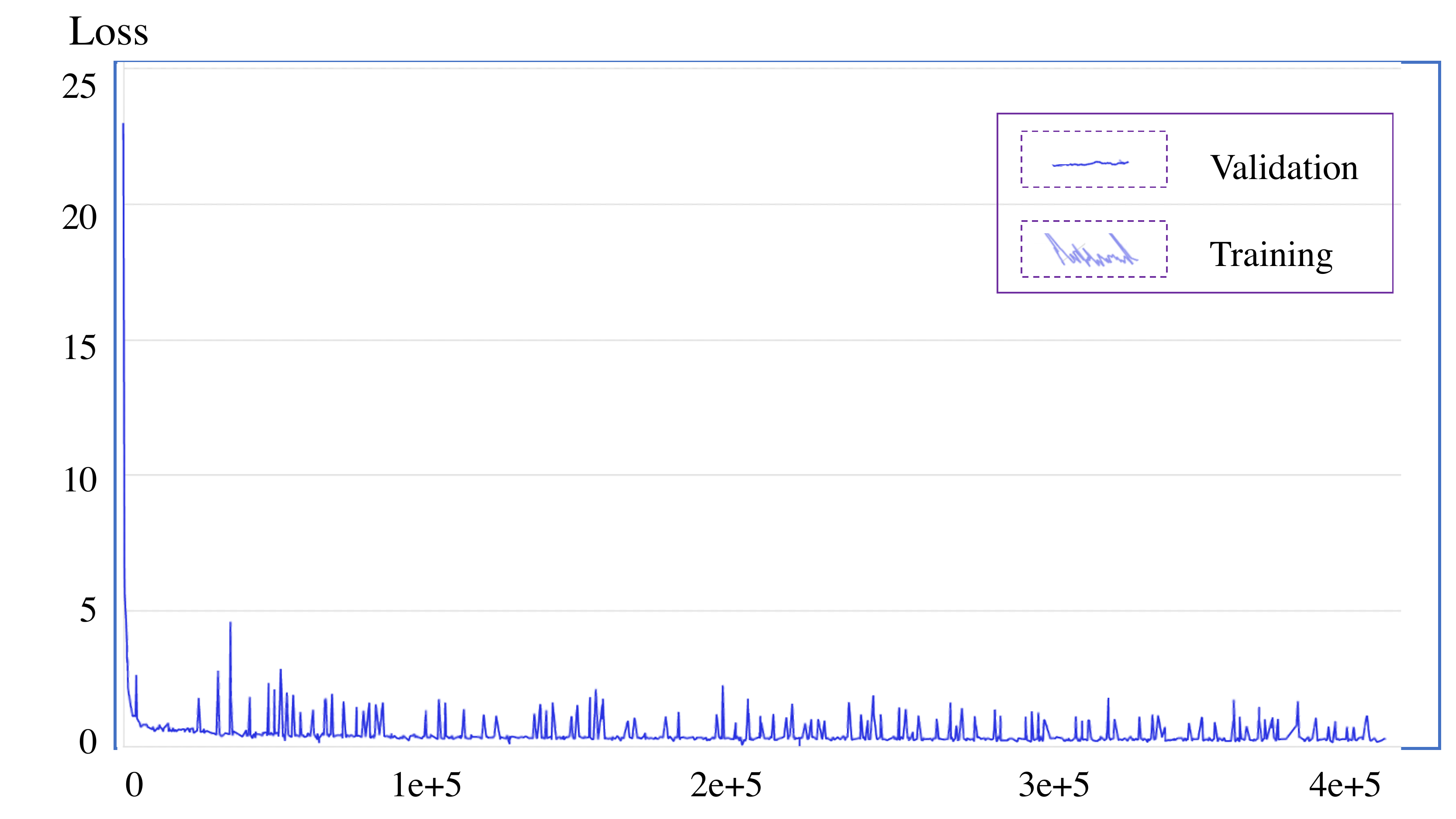}
         \caption{One to many mapping model training in the stage 2.1 of \pjds~}
         \label{fig:loss_model1_one2one}
     \end{subfigure}
     \hfill
     \begin{subfigure}{0.5\textwidth}
         \centering
         \includegraphics[width=\textwidth]{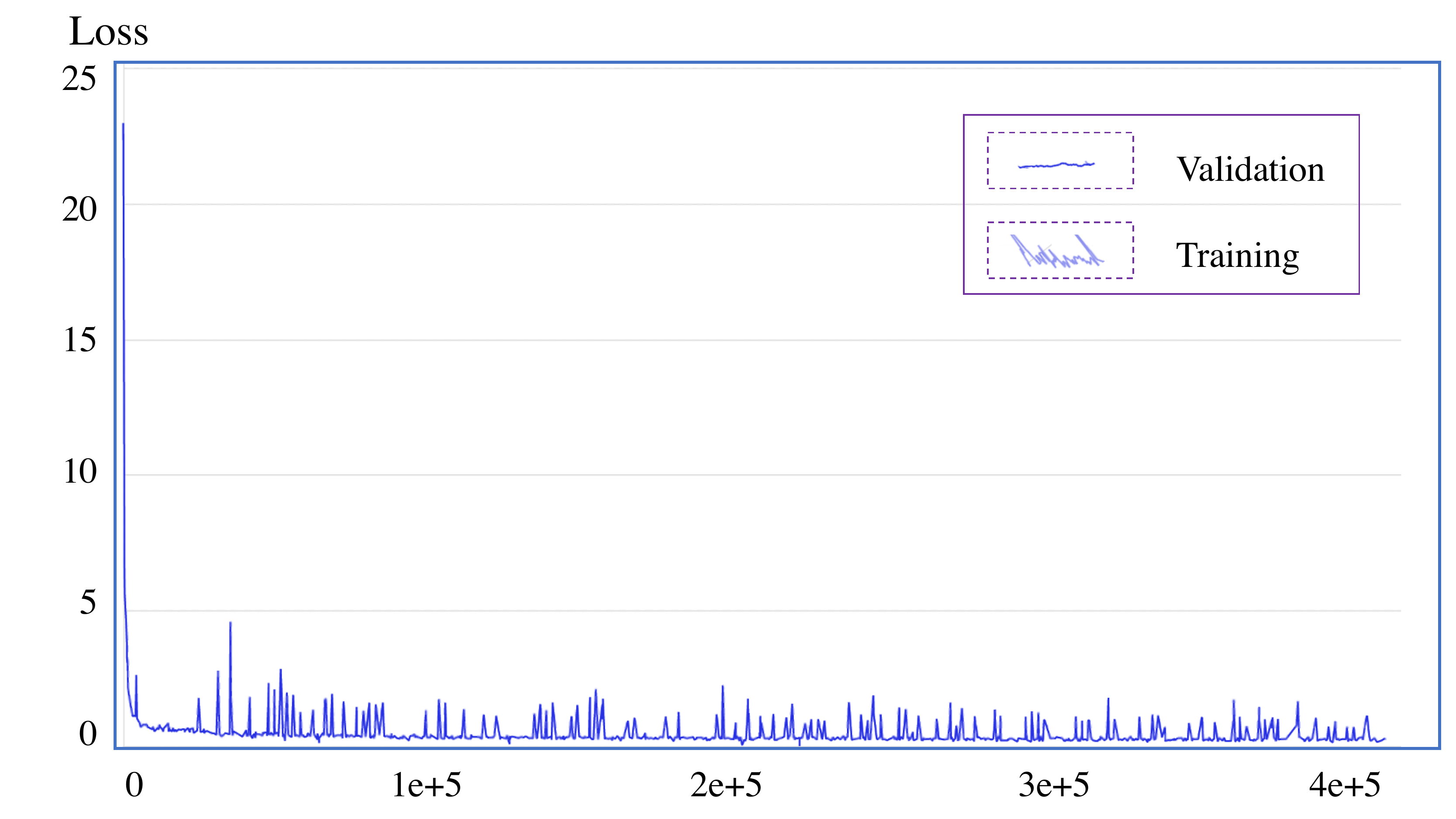}
         \caption{Response evaluation model training in the stage 2.2 of \pjds~}
         \label{fig:loss_model4_one2one}
     \end{subfigure}
        \caption{Three loss graphs of \pjds~ model training in the stage-1, stage-2.1, and stage-2.2.}
        \label{fig:lossgraph}
\end{figure}

\begin{figure}[t!]
 \centering
 \includegraphics[width=0.48\textwidth]{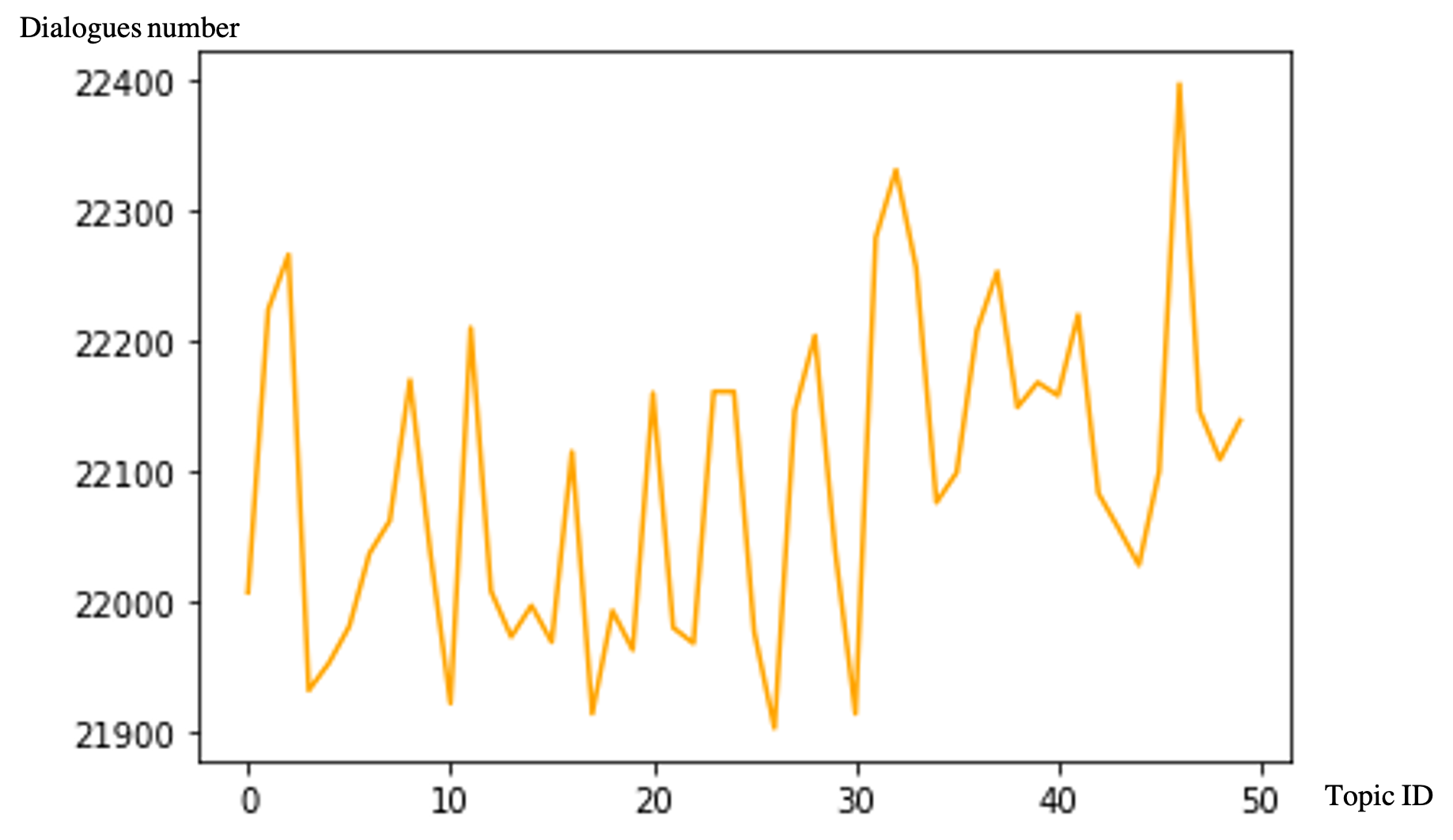}
\caption{The number of dialogues with the change of topics.}
 \label{fig:topic_num}
\end{figure}

Table~\ref{tab:50topics} list all the 50 topics in our data collection.
Table~\ref{tab:start_utterance} list the start utterance of each topic.
Fig.~\ref{fig:topic_num} draw the number of dialogue with the change of topics.
Table~\ref{tab:data_statistic} displays the statistic of our Dialogue-Graph dataset.
Table~\ref{tab:plato-2-english} shows the comparison result of bot-bot chat of PLATO-2 on English data. Table~\ref{tab:twomodel} displays the details of PLATO-JDS training setting.
Fig.~\ref{fig:lossgraph} draws the loss of PLATO-JDS during the training.

\begin{table}[t]
\centering
\begin{adjustbox}{width=0.45\textwidth}
\begin{tabular}{l|c|l|c} 
\thickhline
\textbf{話題}       & \textbf{Topic}            & \textbf{話題}& \textbf{Topic}  \\ \hline
美容と健康(食生活)  & Beauty\&health (eat habits) & 映画       & movie \\ \hline
自己紹介            & self-introduction         & 家族         & family \\ \hline
アウトドア          & outdoors                  & 恋愛         & love \\ \hline
休日の過ごし方      & way to spend holidays     & 趣味         & hobby \\ \hline
友人関係            & friendship                & スポーツ     & do sports \\ \hline
スポーツ            & watch sports              & 勉強         & study \\ \hline
習い事              & lesson                    & 学生生活     & student life \\ \hline
アルバイト          & part-time job             & 料理         & cooking \\ \hline
スイーツ            & sweets                    & お酒         & alcohol \\ \hline
テレビ番組          & TV program                & 通勤         & commute \\ \hline
インテリア          & indoor setting            & 家事        & housework\\ \hline
車、ドライブ        & car, drive                & 仕事         & work\\ \hline
好きな動物          & Favorite animal           & 季節         & season    \\ \hline
家電                & Home appliances           & 携帯電話     & cellphone \\ \hline
旅行(国内)          & Travel (domestic)         & 出身地       & birthplace \\ \hline
美容と健康(運動)    & Beauty\&health (exercise) & 語学         & language \\ \hline
マッサージ          & message                   & カフェ       & Cafe \\ \hline
読書・マンガ        & Reading / manga           & 音楽         & music \\ \hline
欲しいもの          & what I want               & 結婚         & marriage \\ \hline
旅行(海外)          & travel (overseas)         & 食べ物       & food\\ \hline
どこでもドア        & anywhere door             & ランチ       & lunch \\ \hline
好きな芸能人        & favorite entertainer      & 好きな街     & favorite town \\ \hline
ハマっているもの    & addicted things           & 将来の夢     & future dream \\ \hline
ファッション        & fashion                   & 宝くじ       & lottery \\ \hline
睡眠(生活習慣？）   & Sleep (lifestyle?)        & SNS          & SNS \\ \hline
\thickhline
\end{tabular}
\end{adjustbox}
\caption{50 topics list in English and Japanese}
\label{tab:50topics}
\end{table}

\begin{table}[h!]
\centering
\begin{adjustbox}{width=0.45\textwidth}
\begin{tabular}{c|l} 
\thickhline
 \textbf{Topic}         & \textbf{start utterance}  \\ \hline
family                  & Are you living alone? \\ \hline
marriage                & Are you married?? \\ \hline
friendship              & Do you meet friends in school? \\ \hline
hobby                   & What do you do for fun?? \\ \hline
do sports               & Do you do some sports? \\ \hline
study                   & Do you have any studying now? \\ \hline
cooking                 & Do you like cooking? \\ \hline
sweets                  & Do you like sweets? \\ \hline
\multicolumn{2}{c}{......} \\ \hline
\thickhline
\end{tabular}
\end{adjustbox}
\caption{Example of topics and the corresponding start utterances.}
\label{tab:start_utterance}
\end{table}

\begin{table}[t!]
\centering
\begin{adjustbox}{width=0.45\textwidth}
\begin{tabular}{c|c} 
\thickhline
\textbf{\#}         & \textbf{Number of \#}  \\ \hline
Utterance with max number of tokens & 95 \\ \hline
Utterance with min number of tokens & 1 \\ \hline
Utterance with avg. number of tokens & 12.28 \\ \hline \hline
Dialogue with max number of tokens & 258 \\ \hline
Dialogue with min number of tokens & 23 \\ \hline
Dialogue with avg. number of tokens & 109.3 \\ \hline \hline
Number of Turns & 8 \\ \hline \hline
Total dialogues & 1.656M \\ \hline 
Total utterances & 9,837,742 \\ \hline \hline
Max number of all dialogues for a topic  & 22,397 \\ \hline
Min number of all dialogues for a topic  & 21,903 \\ \hline
Avg. number of all dialogues across topics & 22,092.04 \\ \hline 
Variance of all dialogues across topics   & 14106.32 \\ \hline \hline
\thickhline
\end{tabular}
\end{adjustbox}
\caption{Statistics of our Dialogue-Graph dataset.}
\label{tab:data_statistic}
\end{table}

\begin{table}[h]
\centering
\resizebox{1.0\columnwidth}{!}{
\begin{tabular}{c|ccc|ccccc}
\thickhline
\multirow{2}{*}{\textbf{Model\-(Stage)}} & \multicolumn{3}{|c|}{\textbf{Hyper-parameters}} &\multirow{2}{*}{\textbf{PPL}}&\multirow{2}{*}{\textbf{Loss}} &\multirow{2}{*}{\textbf{NLL/RCE loss}} &\multirow{2}{*}{\textbf{Bow loss}} &\multirow{2}{*}{\textbf{MLM loss}} \\
\cline{2-4}
&Batch size & Learning rate &Steps& & & & & \\  \hline
Stage 1& 2,700 &4e-4  &16M  &1.92 &0.52 &0.52 &- &-\\
Stage 2.1&5,400 &4e-5  &1.10M  &1.20 &4.40 &0.23 &4.17 &-\\
Stage 2.2&2,900 &4e-5 &2.54M  &- &0.58 &0.39 &- &0.19\\
\thickhline
\end{tabular}}
\caption{Details of \pjds~training setting. The $NLL$ loss is computed in stage 1, the $NLL$ and $Bow$ losses are calculated in stage 2.1, and the $RCE$ and $MLM$ losses are obtained in stage 2.2 } 
\label{tab:twomodel}
\end{table}

\begin{table}[h]
\centering
\begin{adjustbox}{width=0.45\textwidth}
\begin{tabular}{r|cccc} 
\thickhline
\diagbox[width=7em]{\textbf{Metric}}{\textbf{Model}} & \textbf{PLATO} & \textbf{DialoGPT} & \textbf{Blender} & \textbf{PLATO-2} \\ \hline
\multicolumn{5}{c}{\textbf{Human Evaluation}} \\ \hline
Coherence & 0.568 & 0.720 & 1.856 & \textbf{1.920} \\
Informativeness & 0.564 & 0.712 & 1.816 & \textbf{1.892} \\
Engagingness & 0.340 & 0.340  & 1.820& \textbf{1.840} \\
Humanness & 0.280 & 0.100 & 1.540 & \textbf{1.740}\\ \hline
\multicolumn{5}{c}{\textbf{Automatic Evaluation}} \\ \hline
Distinct-1/2 & 0.042/0.255 & 0.150/0.508 & 0.117/0.385 & \textbf{0.169/0.613} \\ \hline 
\thickhline
\end{tabular}
\end{adjustbox}
\caption{Comparison results of bot-bot chat strategy of PLATO-2 on English data. A higher value indicates higher quality. All the results come from original PLATO-2 paper (Bao et al., 2021a)}
\label{tab:plato-2-english}
\end{table}

\end{document}